\documentclass[10pt,twocolumn,letterpaper]{article}

\usepackage{3dv}
\usepackage{times}
\usepackage{epsfig}
\usepackage{graphicx}
\usepackage{amsmath}
\usepackage{amssymb}
\usepackage{flushend}
\usepackage{float}
\usepackage{algorithm}
\usepackage{algorithmic}
\usepackage{booktabs}

\threedvfinalcopy 

\def\threedvPaperID{****} 

\ifthreedvfinal\fi
\begin{document}

\title{
On 3D Face Reconstruction via Cascaded Regression in Shape Space
}

\author{Feng Liu,~~Dan Zeng,~~Jing Li,~~Qijun Zhao\\
College of Computer Science, Sichuan University, Chengdu, China\\
{\tt\small qjzhao@scu.edu.cn}
\and
\\
}

\maketitle

\begin{abstract}
Cascaded regression has been recently applied to reconstructing 3D faces from single 2D images directly in shape space, and achieved state-of-the-art performance. This paper investigates thoroughly such cascaded regression based 3D face reconstruction approaches from four perspectives that are not well studied yet: (i) The impact of the number of 2D landmarks; (ii) the impact of the number of 3D vertices; (iii) the way of using standalone automated landmark detection methods; and (iv) the convergence property. To answer these questions, a simplified cascaded regression based 3D face reconstruction method is devised, which can be integrated with standalone automated landmark detection methods and reconstruct 3D face shapes that have the same pose and expression as the input face images, rather than normalized pose and expression. Moreover, an effective training method is proposed by disturbing the automatically detected landmarks. Comprehensive evaluation experiments have been done with comparison to other 3D face reconstruction methods. The results not only deepen the understanding of cascaded regression based 3D face reconstruction approaches, but also prove the effectiveness of proposed method.
\end{abstract}

\section{Introduction}
As a fundamental problem in computer vision, reconstructing three dimensional (3D) face shapes from two dimensional (2D) images has recently gained increasing attention because of the 3D face provides invariant features to variations of pose, illumination, and expression. The reconstructed 3D faces are therefore useful for many real-world applications, for example, pose robust face recognition~\cite{blanz2003face, han20123d, hu2014discriminative, zhu2015high}, 3D facial expression analysis~\cite{chu20143d, ren2016face} and facial animation~\cite{cao2014displaced, cao2016real}. Using 3D face shape to recognize identities is believed to be more robust and more accurate than using only 2D face images~\cite{abiantun2014sparse}. Despite its high recognition accuracy, fast acquisition of high resolution and high precision 3D face shapes is still difficult, especially under varying conditions or at a distance. On the other hand, 2D face images can be much more easily captured with the widespread cameras, and there are already plenty of 2D face image databases. It is thus highly demanded to develop efficient methods for reconstructing 3D faces from 2D face images such that the rich resources of 2D face images and facilities can be better utilized. 


A novel method \cite{liu2016joint} has recently been proposed for reconstructing 3D face shapes from single 2D images via cascaded regression in 2D/3D shape space. It is based on the observation that the landmarks' locations on the 2D image can be derived from the reconstructed 3D shape, and the displacement of derived landmarks from their true positions is correlated with the accuracy of the reconstructed 3D shape. This method can simultaneously locate facial landmarks and reconstruct 3D face shapes with two sets of cascaded regressors, one for updating landmarks and the other for 3D face shapes. By effectively exploring the correlation between 2D landmarks and 3D shapes, this method achieves state-of-the-art performance in both face alignment and 3D face reconstruction for face images of arbitrary view and expression. Some problems are, however, still not well answered with regard to such shape space regression based 3D face reconstruction methods:
\begin{itemize}
\begin{item} \emph{Impact of the number of 2D landmarks.} \end{item} Different sets of 2D landmarks are used in the face alignment and recognition literature, e.g., 68 landmarks \cite{Sagonas2013300}, 21 landmarks \cite{K2011Annotated} and 5 landmarks \cite{sun2013deep}. How will the 3D face reconstruction accuracy be affected if different numbers of 2D landmarks are used to guide the 3D face reconstruction process? 
\begin{item} \emph{Impact of the number of 3D vertices.} \end{item} 3D face shapes can be represented by different numbers of vertices, i.e., different 3D point cloud densities and coverage. Will a sparse or narrow 3D face shape be more easily to be reconstructed with higher accuracy than a dense or wide 3D face shape \footnote{A wide 3D face shape covers more areas than a narrow 3D face shape. For instance, the 3D face shape covering only eyes, eyebrows, nose and mouth is narrow compared with the 3D face shape covering the area from left ear to right ear.}?
\begin{item} \emph{What if using standalone landmark localization methods?} \end{item} Although the method in \cite{liu2016joint} can simultaneously locate 2D landmarks and reconstruct 3D shapes, it requires that the training 2D face images should be annotated with both visible and invisible landmarks. Manually marking invisible landmarks is, however, very difficult and error-prone. Is it possible to integrate standalone landmark localization methods with the 3D face reconstruction process proposed in \cite{liu2016joint}?
\begin{item} \emph{Convergence.} \end{item} As an iterative approach, how many iterations would be necessary for the proposed method to achieve acceptable performance in terms of both accuracy and efficiency? In other words, what is the convergence property of shape space regression based 3D face reconstruction methods?
\end{itemize}

The goal of this paper is to investigate the shape space regression based 3D face reconstruction approach from the aforementioned four aspects. To this end, we first revise and implement the method in \cite{liu2016joint} so that the 3D face reconstruction process can take 2D landmarks that are provided by a third party as input, and reconstruct 3D face shapes that have the same pose and expression as the input images, rather than frontal pose and neutral expression
. See Fig. 1 for the results of the method on some photos from the AFW database~\cite{zhu2012face} using the ground truth visible 2D landmarks as input. We then experimentally evaluate the convergence and computational complexity of the implemented 3D face reconstruction method. Afterwards, we conduct extensive experiments to assess the impact of the number of 2D landmarks and the number of 3D vertices on reconstruction accuracy. We finally make an attempt to integrate state-of-the-art landmark localization methods to the 3D face reconstruction process. 

\begin{figure}[t]
\centering{\includegraphics[width=82mm]{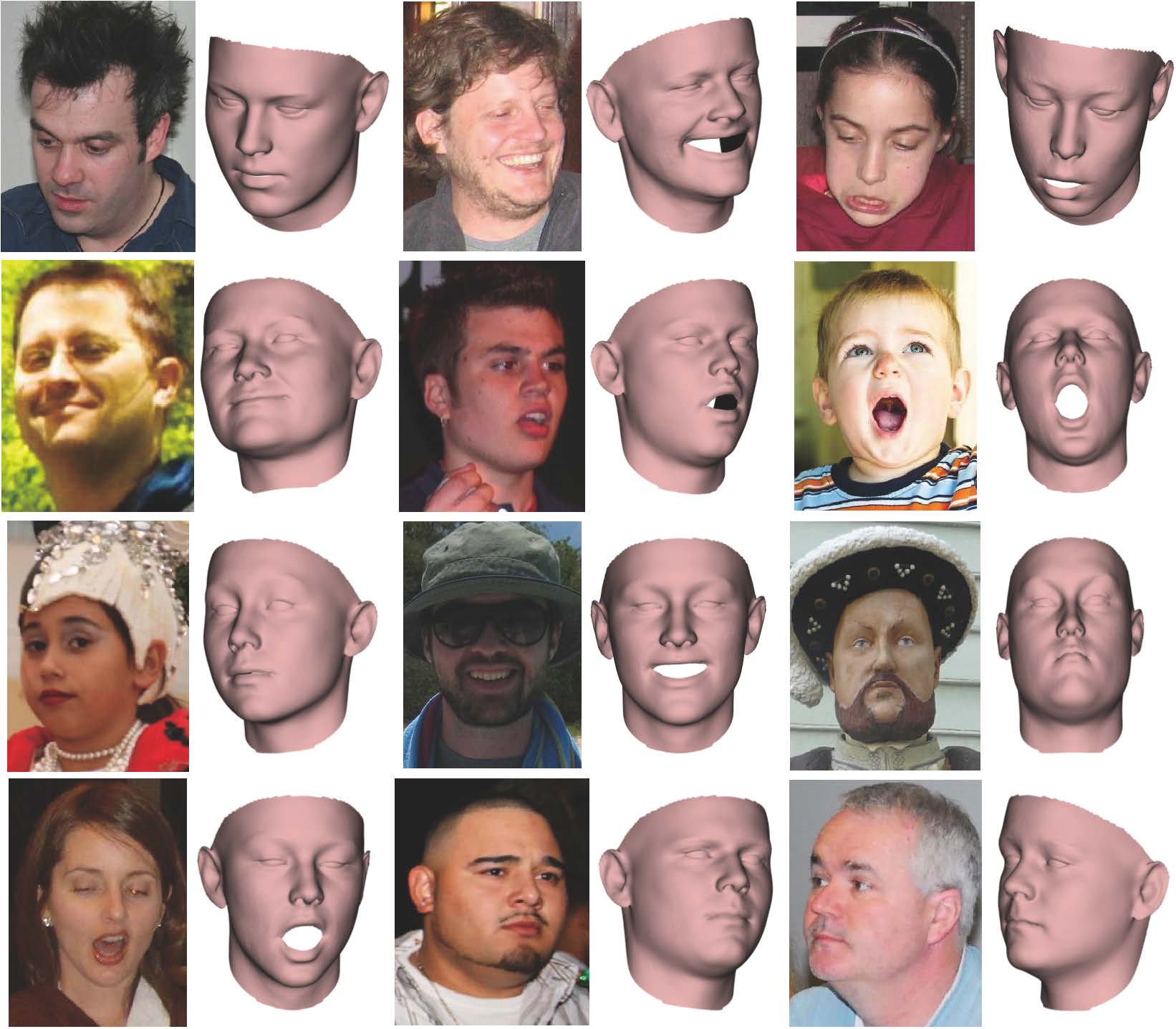}}
\caption{Reconstruction results of the proposed method on face images from the AFW database~\cite{zhu2012face} with arbitrary expressions and poses.}
\label{fig:real_world}
\end{figure}

The rest of this paper is organized as follows. Section 2 reviews the related work. Section 3 and Section 4 present in detail the shape space regression based 3D face reconstruction method and its implementation. Section 5 reports the experimental results. Section 6 finally concludes the paper.

\begin{figure}[t]
\centering{\includegraphics[width=82mm]{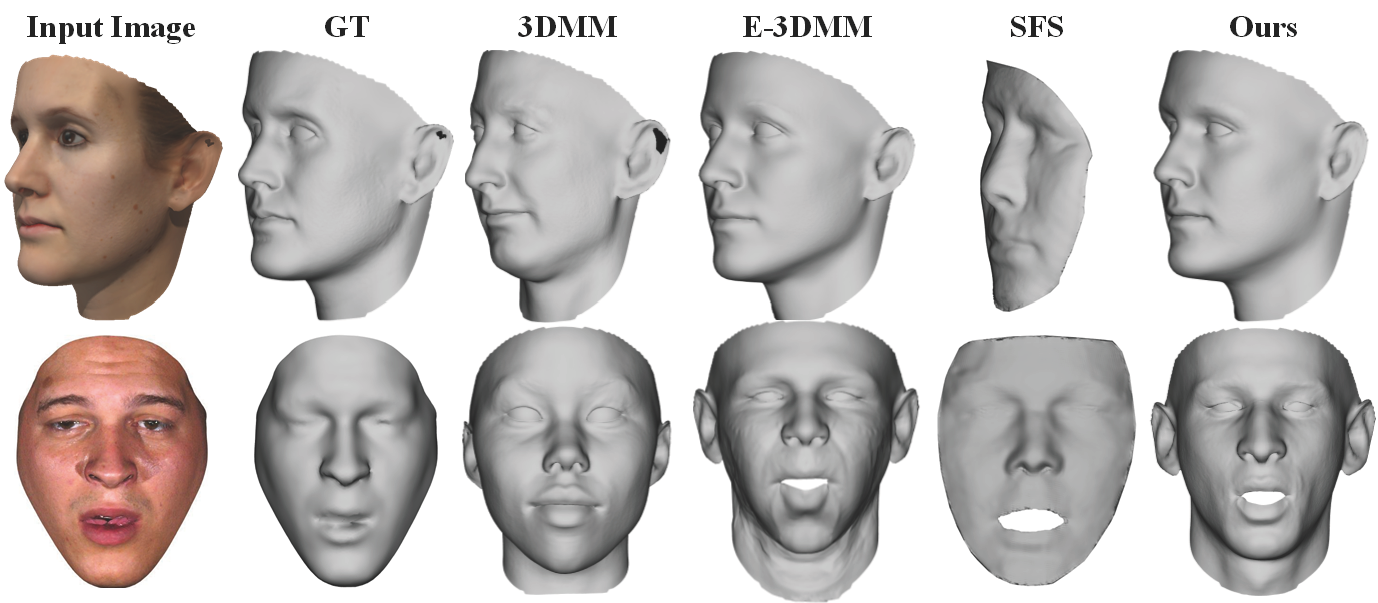}}
\caption{Reconstruction results for images in the Basel Face Model (BFM) (top row)~\cite{paysan20093d} and BU3DFE (bottom row)~\cite{yin3d_2006} databases. From left to right columns: Input images, ground truth 3D shapes (GT), and results by 3DMM~\cite{bas2016fitting}, E-3DMM~\cite{zhu2015high}, SFS~\cite{kemelmacher20113d} and our proposed method.}
\label{fig:method_compare}
\end{figure}

\section{Related Work}
In order to solve the intrinsically ill-posed single-view 3D face reconstruction problem, different priors or constraints have been introduced, resulting in the Shape from Shading (SFS) based methods and 3D Morphable Model (3DMM) based methods. SFS based methods~\cite{horn1989shape, barron2012shape} recover 3D shapes via analyzing certain clues in the 2D texture images, with an assumption of the Lambertian reflectance and a single-point light source at infinity. While classical SFS based methods~\cite{kemelmacher20113d, suwajanakorn2014total, li2015example-based, zeng2017examplar} are initially designed for generic 3D shape reconstruction, their performance in recovering 3D face shapes can be further improved by using some reference 3D face models as additional constraints. These methods usually have limited accuracy because (i) their assumed connection between 2D texture clues and 3D shape information is too weak to discriminate between different human faces, (ii) they do not fully exploit the prior knowledge of 3D faces and significantly depend on the reference models, and (iii) they reconstruct a depth map or 2.5D shape instead of a 3D full shape since they tend to operate on a face with a narrow range of poses.

3D Morphable Model (3DMM)~\cite{blanz1999morphable, romdhani2005estimating, aldrian2013inverse, zhu2014robust, zhu2015high, Booth_2016_CVPR, bas2016fitting}, as a typical statistical 3D face model, explicitly learns the prior knowledge of 3D faces with a statistical parametric model. It represents a 3D face as a linear combination of basis 3D faces, which are obtained by applying principal component analysis (PCA) on a set of densely aligned 3D faces. To recover the 3D face from a 2D image, the combination coefficients are estimated by minimizing the discrepancy between the input 2D face image and the one rendered from the reconstructed 3D face. These 3DMM based methods can better cope with 2D images of varying illuminations and poses. However, they are limited in individualized or detail reconstruction because PCA conducts global modeling in essence, and they involve a time-consuming on-line optimization process to search for optimal solution in the parameter space. Moreover, 3DMM needs an additional linear expression model to handle facial expressions, namely E-3DMM~\cite{chu20143d, cao2014facewarehouse, zhu2015high}. However, neither SFS-based nor 3DMM-based methods can consistently well cope with rotated or expressive face images due to invisible or deformed facial landmarks on them.

Motivated by the success of cascaded regression in 2D facial landmark localization \cite{xiong2013supervised, jourabloo2015pose, li2016affine}, the authors recently proposed in \cite{liu2016joint} a 2D/3D shape space regression based method for reconstructing 3D face shapes from single images of arbitrary views and expressions. The method alternately applies 2D landmark regressors and 3D shape regressors. The 2D landmark regressors estimate landmark locations by regressing over the texture features around landmarks, while the 3D shape regressors reconstruct 3D face shapes via regressing over the 2D landmarks. Unlike existing 3D face reconstruction methods, this method directly estimates 3D faces in the 3D shape space via cascaded regression, getting rid of parameterized 3D face models and assumed illumination models. As a result, it achieves state-of-the-art performance for both accuracy and efficiency of 3D face reconstruction. Figure~\ref{fig:method_compare} shows example results of SFS-based, 3DMM-based, E-3DMM-based and shape-space-regression-based methods on rotated and expressive face images. In this paper, we will thoroughly assess the effectiveness of such shape space regression based 3D face reconstruction methods from various perspectives.



\section{Shape Space Regression based Approach}
\subsection{Overview} 
We denote a 3D face shape as $\mathbf{S} \in \mathbb{R}^{3\times n}$, which is represented by 3D locations of $n$ vertices, and a subset of $\mathbf{S}$ with columns corresponding to $l$ annotated landmarks (e.g., eye corners and nose tip) as $\mathbf{S}_{L}$. The projections of these 3D landmarks on the 2D face image $\mathbf{I}$ are represented by $\mathbf{U} \in \mathbb{R}^{2\times l}$. The relationship between 2D facial landmarks $\mathbf{U}$ and its corresponding 3D landmarks $\mathbf{S}_{L}$ can be described as:
\begin{equation}
\label{eqn::camera_projection}
\mathbf{U} = M \mathbf{S}_{L} = M D_{N} ( {R}\tilde{\mathbf{S}}_{L} + {T} ),
\end{equation}
where $\tilde{\mathbf{S}}$ is a frontal 3D face with neutral expression, $\{{R}\in \mathbb{R}^{3\times3}, {T}\in \mathbb{R}^{3\times l}\}$ and $D_{N}(\cdot)$ are, respectively, rigid deformation (i.e., rotation and translation) caused by pose variations and non-rigid deformation function caused by expression variations that occur to $\tilde{\mathbf{S}}$ resulting in the observed 3D face $\mathbf{S}$, and $M\in \mathbb{R}^{2\times3} $ is the camera projection matrix. Here, we employ weak perspective projection for $M$ as conventionally done in the literature \cite{zhou20153d}. 

\begin{figure}[t]
\centering{\includegraphics[width=78mm]{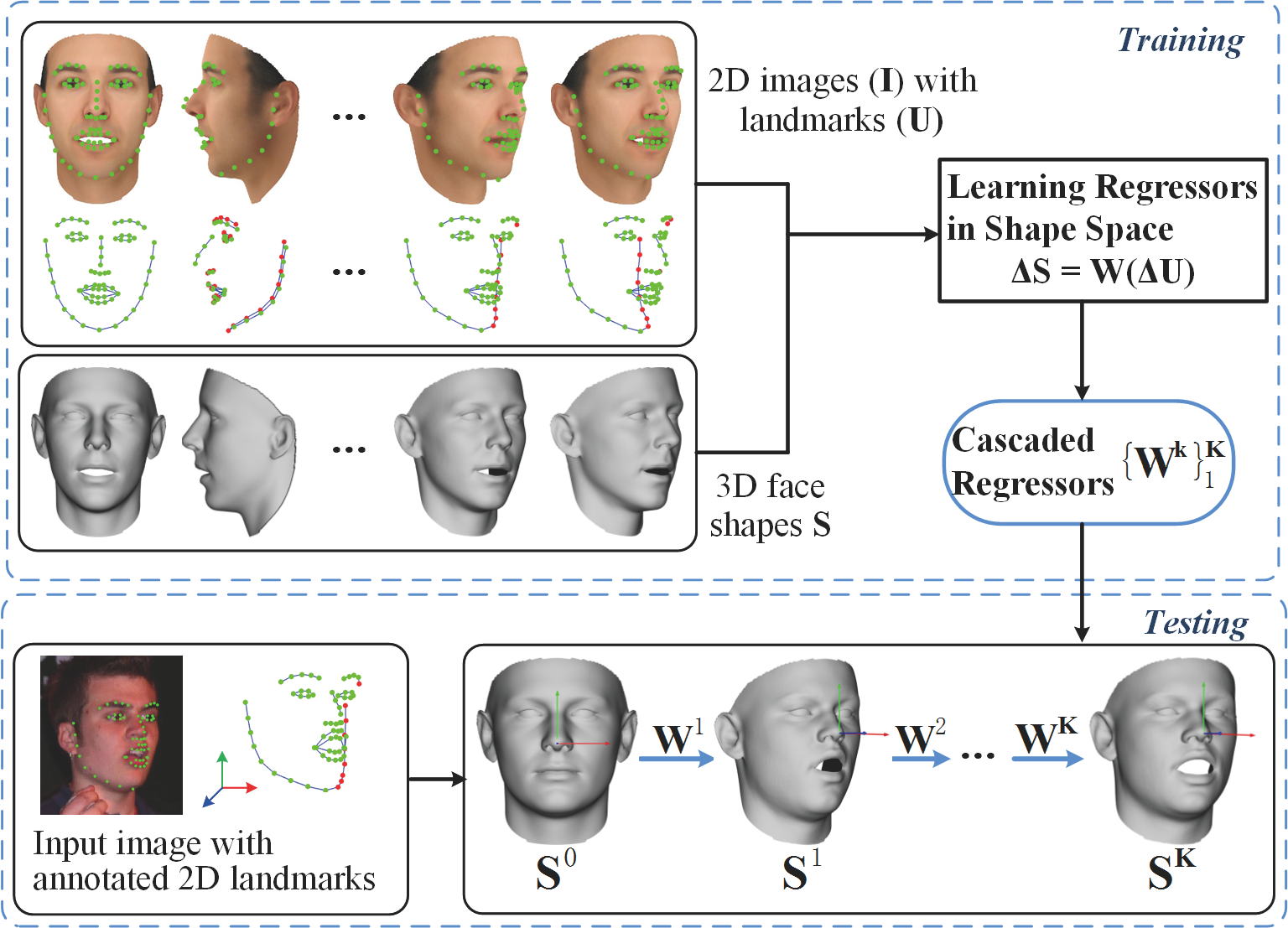}}
\caption{Flowchart of the shape space cascaded regression based 3D face reconstruction method. Green and red points denote, respectively, visible and invisible landmarks. Note that the method in this paper does not require invisible landmarks' locations as input.}
\label{fig:algorithm}
\end{figure}

Our purpose in this paper is to reconstruct $\mathbf{S}$ (rather than $\tilde{\mathbf{S}}$) from the given ``ground truth'' visible landmarks $\mathbf{U}^{*}$ (either manually marked or automatically detected by a standalone method) for the face image $\mathbf{I}$. As discussed above, we achieve this by iteratively updating the initial estimate of $\mathbf{S}$ with a series of regressors in the 3D face shape space. These regressors calculate the adjustment to the estimated 3D face shape according to the deviation between the ground truth landmarks and the landmarks rendered from the estimated 3D face shape.  Figure \ref{fig:algorithm} shows the flowchart of the proposed method.

\subsection{The Reconstruction Process}

Let $\mathbf{U}^{*}$ be the ``ground truth'' landmarks (either manually annotated or automatically detected) on an input 2D image, and $\mathbf{S}^{k-1}$ the currently reconstructed 3D shape after $k-1$ iterations. The corresponding landmarks $\mathbf{U}^{k-1}$ can be obtained by projecting $\mathbf{S}^{k-1}$ onto the image according to Eqn.~(\ref{eqn::camera_projection}). Then the updated 3D shape $\mathbf{S}^{k}$ can be computed by
\begin{equation}
\label{eqn::shape_increment}
\mathbf{S}^{k} = \mathbf{S}^{k-1}+\mathbf{W}^{k}(\mathbf{U}^{*}-\mathbf{U}^{k-1}) + \mathbf{b}^{k},
\end{equation}
where $\mathbf{W}^{k}$ is the regressor in $k^{th}$ iteration and $\mathbf{b}^{k}$ is a bias term (in the rest of this paper we omit the bias term for simplicity sake because it can be shrunk into the regressors). 

\subsection{Learning Cascaded Regressors} \label{sec:learning}
The $K$ regressors $\{\mathbf{W}^{k}\}_{1}^{K}$ involved in the reconstruction process can be learned via optimizing the following objective function over the $N$ training samples:
\begin{equation}
\label{eqn::objectiveFun}
 \mathop {\arg \min }\limits_{{\mathbf{W}^{k}}}\sum_{i=1}^N\parallel(\mathbf{S}_{i}^{*} - \mathbf{S}^{k-1}_{i}) - \mathbf{W}^{k}(\mathbf{U}_{i}^{*}-\mathbf{U}_{i}^{k-1})\parallel_2^2,
\end{equation}
where $\{\mathbf{S}_{i}^{*}, \mathbf{U}_{i}^{*}\}$ is one training sample consisting of ground truth landmarks on the $i^{\texttt{th}}$ 2D face image and its corresponding ground truth 3D face shape that has the same pose and expression as the face image. Mathematically, the above optimization seeks for a regressor that can minimize the overall error of the entire reconstructed 3D face shapes, but not merely the error at the landmarks.

In this paper, we use linear regressors $\mathbf{W}^{k}\in\mathbb{R}^{(3n) \times (2l)}$. The optimization in Eqn.~(\ref{eqn::objectiveFun}) can be then easily solved by using least squares methods with a solution of 
\begin{equation}
\label{eqn::LS}
\mathbf{W}^{k} = \Delta \mathbb{S}^{k}(\Delta \mathbb{U}^{k})^{\mathsf{T}}(\Delta \mathbb{U}^{k}(\Delta \mathbb{U}^{k})^{\mathsf{T}} )^{-1}, 
\end{equation}
where $\Delta \mathbf{\mathbb{S}}^{k} = \mathbf{\mathbb{S}}^{*}-\mathbf{{\mathbb{S}}}^{k-1}$ and $\Delta \mathbf{\mathbb{U}}^{k} = \mathbf{\mathbb{U}}^{*} - \mathbf{\mathbb{U}}^{k-1}$ are 3D shape adjustment and 2D landmark deviation. $\mathbb{S}\in \mathbb{R}^{(3n)\times N}$ and $\mathbb{U}\in \mathbb{R}^{(2l)\times N}$ denote, respectively, the ensemble of 3D face shapes and 2D landmarks of all training samples with each column corresponding to one sample. Note that, here, we write 3D face shape and 2D landmarks as column vectors: $\mathbf{S}=(x_{1}, y_{1}, z_{1}, x_{2}, y_{2}, z_{2}, \cdots, x_{n}, y_{n}, z_{n})^{\mathsf{T}}$ and $\mathbf{U}=(u_{1}, v_{1}, u_{2}, v_{2}, \cdots, u_{l}, v_{l})^{\mathsf{T}}$ (`$\mathsf{T}$' denotes transpose operator). It can be mathematically shown that, to ensure a valid solution in Eqn.~(\ref{eqn::LS}), $N$ should be larger than $2l$ so that $\Delta \mathbb{U}^{k}(\Delta \mathbb{U}^{k})^{\mathsf{T}}$ is invertible. Fortunately, since the set of used landmarks are usually sparse, this requirement can be easily satisfied in real-world applications. 

\begin{algorithm}
\caption{3D Cascaded Regressor Learning}
\label{alg:reconstruction_algorithm}
\begin{algorithmic}[1]
\REQUIRE Training data $\{(\mathbf{I}_{i}, \mathbf{S}^{*}_{i}, \mathbf{U}^{*}_{i})\vert i=1,2,\cdots,N\}$, initial shape $\mathbf{S}^{0}_{i}$ \& camera projection matrix $\mathbf{M}$.
\ENSURE Cascaded regressors $ \{\mathbf{W}^{k} \}_{k=1}^{K}$.
\FOR{$k=1,...,K$} 
\STATE Estimate 2D projection $\mathbf{U}^{k-1}_{i}$ from current 3D face $\mathbf{S}^{k-1}_{i}$ via Eq. (\ref{eqn::camera_projection}); 
\STATE Compute 2D landmark adjustment and 3D face adjustment for all samples: $\Delta \mathbf{\mathbb{U}}^{k} = \mathbf{\mathbb{U}}^{*} - \mathbf{\mathbb{U}}^{k-1}$, $\Delta \mathbf{\mathbb{S}}^{k} = \mathbf{\mathbb{S}}^{*}-\mathbf{{\mathbb{S}}}^{k-1}$;
\STATE Estimate $\mathbf{W}^{k}$ via Eqn.~(\ref{eqn::objectiveFun});
\STATE Update 3D face $\mathbf{S}^{k}_{i}$ via Eqn.~(\ref{eqn::shape_increment}).
\ENDFOR 
\end{algorithmic}
\end{algorithm}

\section{Implementation Details}

\subsection{Initialization}
The proposed iterative method has two terms to initialize: the initial 3D face shape $\mathbf{S}^{0}$ and the camera projection matrix $\mathbf{M}$. Given the set of training samples, we select out from them all the frontal faces with neutral expression. The mean of these selected 3D face shapes is computed and used to initialize $\mathbf{S}^{0}$. Similarly, the mean of their 2D landmarks is also calculated and denoted as $\mathbf{U}^{0}$. The camera projection matrix $M$ can be then estimated by solving the following least squares fitting problem:
\begin{equation}
\label{eqn::compute_m}
\mathbf{M} = \mathop {\arg \min } \limits_{\mathbf{M}} \parallel \mathbf{U}^{0} - \mathbf{M}\times \mathbf{S}^{0}_{L}\parallel_2^2.
\end{equation}
The obtained projection matrix $\mathbf{M}$ is used throughout the 3D face reconstruction process to render 2D landmarks from the reconstructed 3D face shapes.

\begin{figure}[t]
\centering{\includegraphics[width=80mm]{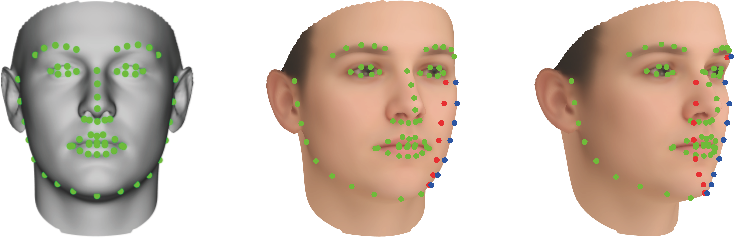}}
\caption{Sixty-eight landmarks are used in this work. Left: Landmarks annotated on a 3D face. Middle and Right: Corresponding landmarks annotated on its 2D images with yaw angle of $20^{\circ}$ and $40^{\circ}$. Green and red points on the 2D images indicate, respectively, visible and invisibile landmarks, and blue points mark the contour instead of semantic landmarks.}
\label{fig:invisible_landmarks}
\end{figure}

\subsection{Landmarks}\label{sec:landmarks}
Figure \ref{fig:invisible_landmarks} depicts the sixty-eight facial landmarks $(l=68)$ considered in this paper. Obviously, some of the landmarks will become invisible on the 2D face images due to self-occlusion when the face has large pose angles. These invisible landmarks are difficult to be precisely annotated. Hence, we treat them as missing data, and fill their corresponding entries in $\mathbf{U}$ with zero. This way, these invisible landmarks will not affect the reconstruction, and thus images of arbitrary pose angles can be handled in a unified framework.

To automatically detect the visible landmarks in testing phase, we first employ state-of-the-art face alignment approach to automatically locate 2D landmarks positions, and then compute their visibility. Most conventional face alignment methods like~\cite{kazemi2014one} can not detect invisible self-occluded landmarks (refer to the red point in Fig. \ref{fig:invisible_landmarks}). In order to determine the visibility of 2D landmarks projected from the reconstructed 3D face shape, given the detected 2D landmarks $\mathbf{U}$ on the face image and the 3D annotated landmarks $\mathbf{S}_{L}^{0}$ from the initial 3D shape $\mathbf{S}^{0}$, we coarsely estimate the camera projection matrix $\mathbf{M}$ by Eqn.~(\ref{eqn::compute_m}). Suppose the 3D surface normal at landmarks in $\mathbf{S}^{0}$ is $\vec{\mathbf{N}}$. The initial visibility $\mathbf{v}$ can be then measured by \cite{Jourabloo_2016_CVPR}

\begin{equation}
\label{eqn::visibility}
\textbf{v} = \frac{1}{2}\left(1 + sgn\left(\vec{\mathbf{N}} \cdot \left( \frac{\mathbf{M}_{1}}{\left \| \mathbf{M}_{1} \right \|}\times \frac{\mathbf{M}_{2}}{\left \| \mathbf{M}_{2} \right \|} \right)\right)\right),
\end{equation}
where $sgn()$ is the sign function, `$\cdot$' means dot product and `$\times$' cross-product, and $\mathbf{M}_{1}$ and $\mathbf{M}_{2}$ are the left-most three elements at the first and second row of the mapping matrix $\mathbf{M}$. This basically rotates the surface normal and validates if it points toward the camera or not. Finally, to maintain the consistence with the training setting, the invisible corresponding entries in $\mathbf{U}$ should be filled with zero.

\subsection{Alignment}
For the sake of simplifying the camera projection model, we assume that both 3D face shapes and 2D landmarks are well aligned. More specifically, (i) all the 3D face shapes have been established point-to-point dense registration (i.e., they have the same number of vertices, and the vertices of the same index have the same semantic meaning); (ii) all the 3D face shapes are centered at the origin of the world coordinate system; and (iii) all the faces on the 2D images are also centered in the image coordinate system. With these aligned 3D$\&$2D face data, and as we separate face deformation from camera projection (see Eqn.~(\ref{eqn::camera_projection})), the employed weak perspective camera projection matrix $M$ has only one free parameter, i.e., scaling factor or focal length, which will be estimated based on the training data.

\begin{figure*}[t]
\centering{\includegraphics[width=164mm]{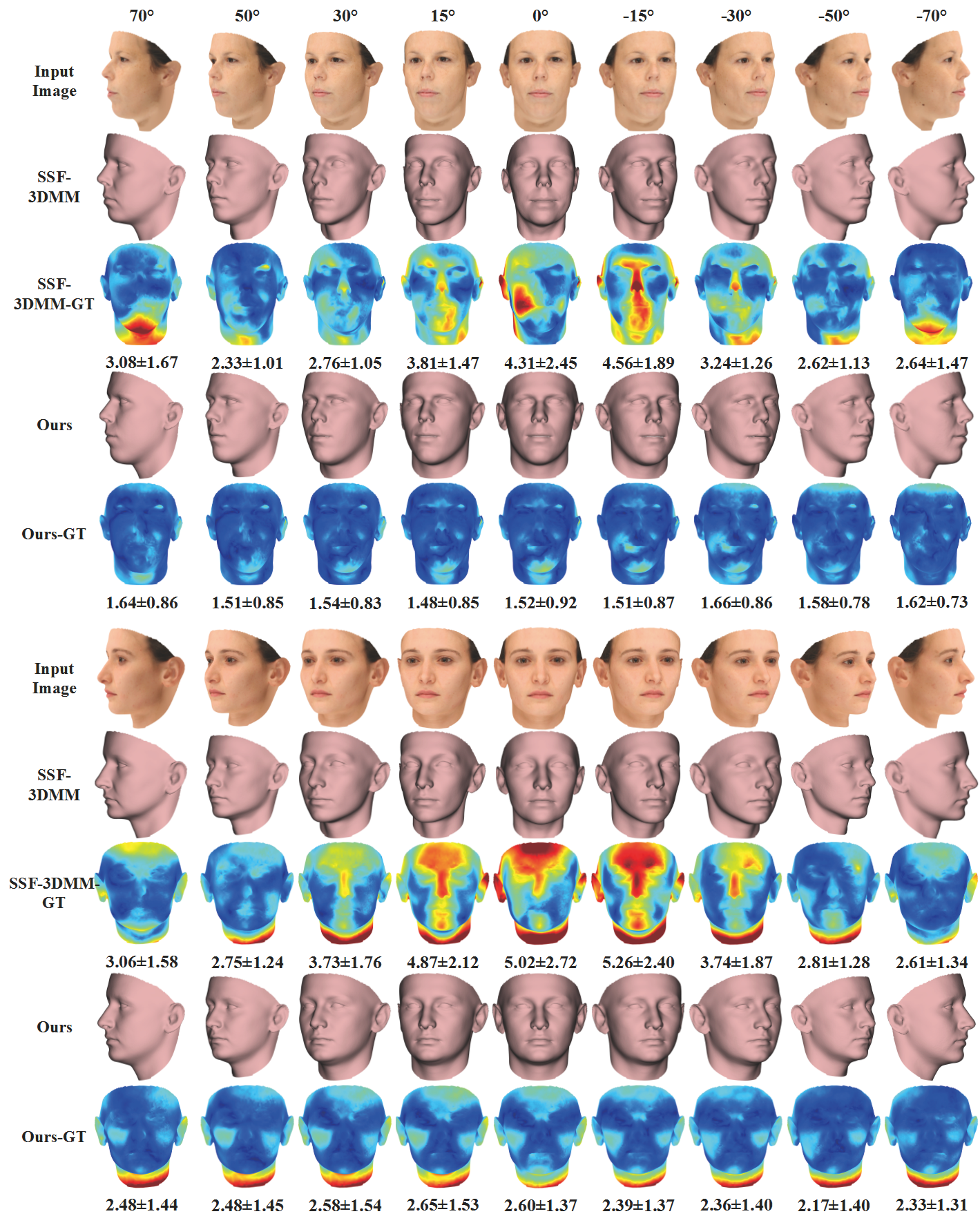}}
\caption{Reconstruction results for two BFM samples at 9 different poses. First row: The input images. Second and forth rows: The reconstructed 3D face shapes by the method of SSF-3DMM~\cite{zhu2014robust} and our proposed method. Third and fifth rows: Their corresponding MAE error maps. The colormap goes from dark blue to dark red (corresponding to an error between $0$ and $10$). The numbers under each of the error maps represent mean and standard deviation values ($mm$).}
\label{fig:bfm_example}
\end{figure*}

\section{Experimental Results}
\subsection{Training Data}
A set of 3D face shapes and corresponding 2D face images with annotated landmarks is needed to train regressors in the proposed method. To make the trained regressors robust to pose and expression variations, samples in the training dataset should have good diversity in their poses and expressions. It is, however, difficult to find in the public domain such datasets of 3D face shapes and corresponding annotated 2D images with various expressions/poses. Therefore, we use the Basel Face Model (BFM)~\cite{paysan20093d} to construct synthetic 3D faces of 200 subjects (50\% female), and use the expression model from FaceWarehouse \cite{cao2014facewarehouse} to generate random expressions on each of the 3D faces. These expressive 3D faces are then projected onto 2D images with 55 views of 11 yaw ($0^{\circ}$, $\pm15^{\circ}$, $\pm30^{\circ}$, $\pm50^{\circ}$, $\pm70^{\circ}$, $\pm90^{\circ}$) and 5 pitch ($0^{\circ}$, $\pm15^{\circ}$, $\pm30^{\circ}$) rotations, resulting in a total number of 11,000 3D faces and corresponding synthetic images. Each 3D face consists of 53,215 vertices (the original BFM model has 53,490 vertices, but we discard the vertices in tongue region). The 2D image resolution is $875 \times 656$ pixels and the inter-eye distance is about $220$ pixels. The 68 landmarks on each 2D face image are recorded during the projection process (note that the 3D faces are densely aligned and the indices of the landmarks in the 3D face shapes are known), and the invisible landmarks are marked as zero as mentioned above.

\begin{table*}[t]
\centering 
\caption{MAE (mm) of the proposed method and four state-of-the-art methods at different poses with ground truth landmarks.} 
\begin{tabular*}{14.2cm}{@{\quad}l@{\quad}c@{\quad}c@{\quad}c@{\quad}c@{\quad}c@{\quad}c@{\quad}c@{\quad}c@{\quad}c@{\quad}c}
\toprule
& \multicolumn{9}{c}{\textbf{Rotation angle}} \\ 
\cmidrule(l){2-10} 
\textbf{Method} & $-70^{\circ}$ & $-50^{\circ}$ & $-30^{\circ}$ & $-15^{\circ}$ & $0^{\circ}$ & $15^{\circ}$ & $30^{\circ}$ & $50^{\circ}$ & $70^{\circ}$ & \textbf{Mean}\\ 
\midrule 
\midrule 
Romdhani et al. \cite{romdhani2005estimating} & 2.65 & 2.59 & 2.58 & 2.61 & 2.59 & 2.50 & 2.50 & 2.46 & 2.51 & 2.55\\ 
Aldrian and Smith \cite{aldrian2013inverse} & 2.64 & 2.60 & 2.58 & 2.64 & 2.56 & 2.49 & 2.50 & 2.54 & 2.63 & 2.58\\ 
SSF-3DMM \cite{zhu2014robust} & 3.45 & 2.81 & 3.71 & 4.62 & 4.97 & 4.81 & 3.74 & 2.98 & 3.19 & 3.81\\ 
Bas et al. \cite{bas2016fitting} & 2.35 &  \textbf{2.26} & 2.38 & 2.40 & 2.51 & 2.39 & 2.40 & \textbf{2.20} &  \textbf{2.26} & 2.35\\ 
\midrule 
Proposed &  \textbf{2.29} & 2.30 &  \textbf{2.35} &  \textbf{2.29} &  \textbf{2.31} &  \textbf{2.27} &  \textbf{2.36} &  2.21 & 2.32 &  \textbf{2.30}\\ 
\bottomrule 
\end{tabular*}
\label{tab:bfm_compare}
\end{table*}

\subsection{Convergence and Computational Complexity}
\begin{figure}[H]
\centering{\includegraphics[width=70mm]{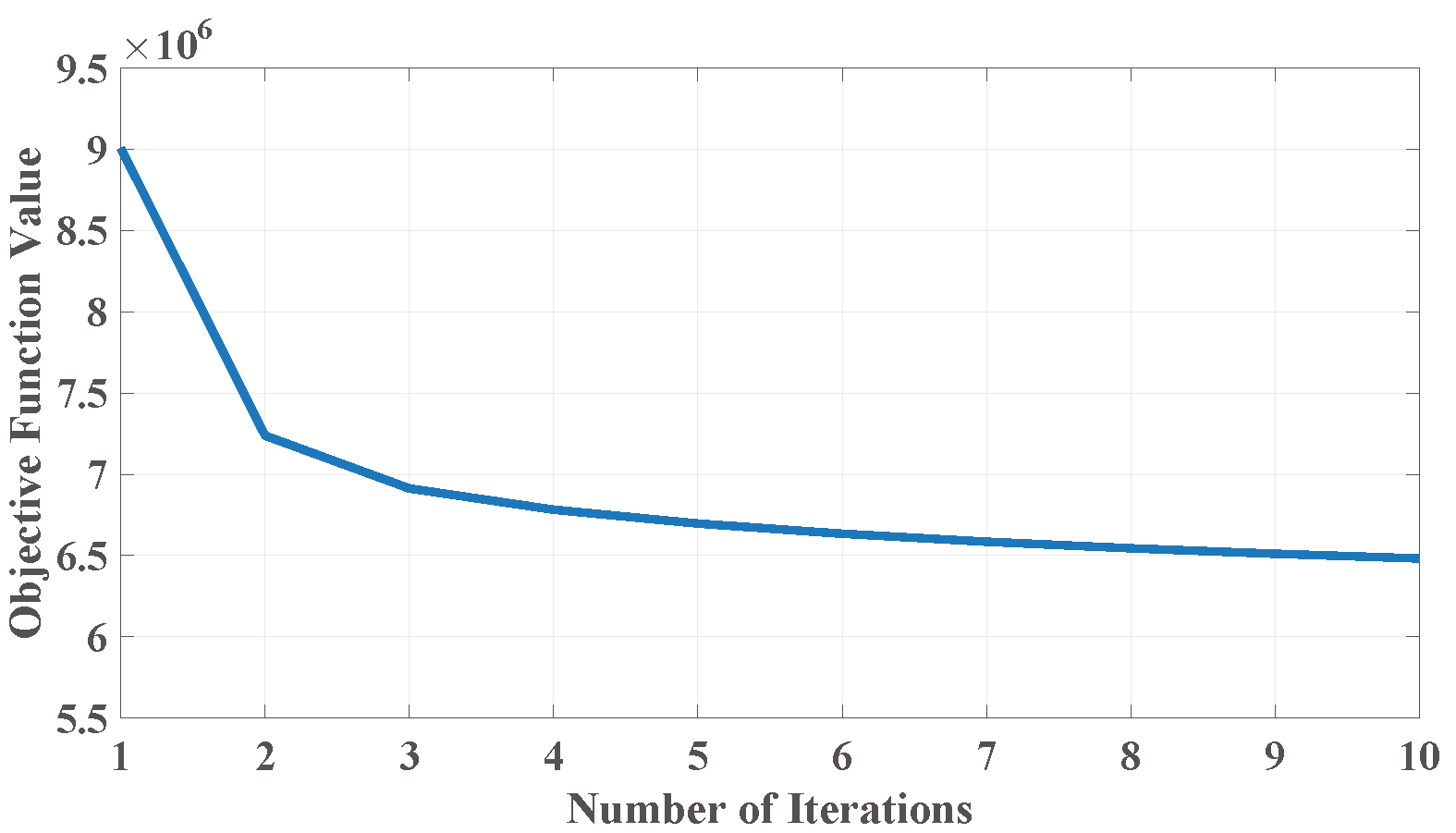}}
\caption{The objective function values as iteration proceeds.}
\label{fig:iteration_error}
\end{figure}

In this section, we experimentally investigate the convergence of the training process of the proposed cascaded regressors. To this aim, we record down the value of the objective function defined in Eqn.~(\ref{eqn::objectiveFun}) at each iteration during the training process. Figure \ref{fig:iteration_error} shows the objective function value for 10 iterations. It can be clearly seen that the objective function value decreases substantially in the first five iterations, and becomes stable after seven iterations. This demonstrates the good convergence of the proposed method. In the following experiments, we empirically set $K=5$ as a trade-off between accuracy and efficiency.

According to our experiments on a PC with i7-4710 CPU and 8 GB memory, the Matlab implementation of the proposed method runs at $\sim26$ frames per second (FPS). This indicates that the proposed method can reconstruct 3D faces in real time.

\subsection{Reconstruction Accuracy across Poses on BFM}

The BFM database~ \cite{paysan20093d} provides 10 test face subjects, each of whom has nine face images of neutral expression and different poses, including one frontal and eight yaw poses ($\pm15^{\circ}$, $\pm30^{\circ}$, $\pm50^{\circ}$, $\pm70^{\circ}$). Here, the metric used to evaluate the 3D face shape reconstruction accuracy is Mean Absolute Error (MAE). MAE is defined as 
\begin{equation}
\label{eqn::MAE}
\texttt{MAE} = \frac{1}{N_{T}}\sum_{i=1}^{N_{T}}(\| \mathbf{S}^{*}_{i}-\hat{\mathbf{S}}_{i} \|/n),
\end{equation}
where $N_{T}$ is the total number of test samples, $\| \mathbf{S}^{*}_{i}-\hat{\mathbf{S}}_{i} \|$ is the Euclidearn distance between ground truth shape $\mathbf{S}^{*}_{i}$ and reconstructed 3D shape $\hat{\mathbf{S}}_{i}$ of the $i^{\texttt{th}}$ test sample. We report the MAE in $mm$ after Procrustes alignment.

In this experiment, we use the visible landmarks projected from ground truth 3D face shapes as input. The proposed method is compared with several state-of-the-art methods based on 3DMM, including the approach proposed by Aldrian and Smith~\cite{aldrian2013inverse}, the multi-features 3DMM framework based on contours, textured edges, specular highlights and pixel intensity proposed by Romdhani et al.~\cite{romdhani2005estimating}, Sparse SIFT Flow 3DMM (SSF-3DMM, ~\cite{zhu2014robust}), and the edge-fitting based 3DMM approach proposed by Bas et al.~\cite{bas2016fitting}.

Table~\ref{tab:bfm_compare} shows the MAE of different methods on the BFM database with respect to different poses of face images. As can be seen, average MAE of the proposed method is obviously lower than that of the counterpart methods. Moreover, its accuracy is very stable across different poses. This proves the effectiveness of the proposed method in handling face images of arbitrary poses. Figure~\ref{fig:bfm_example} shows the reconstruction results of our method and SSF-3DMM~\cite{zhu2014robust} on two subjects in the BFM database.

\subsection{Impact of the Number of 2D Landmarks}
\begin{figure}[h]
\centering{\includegraphics[width=78mm]{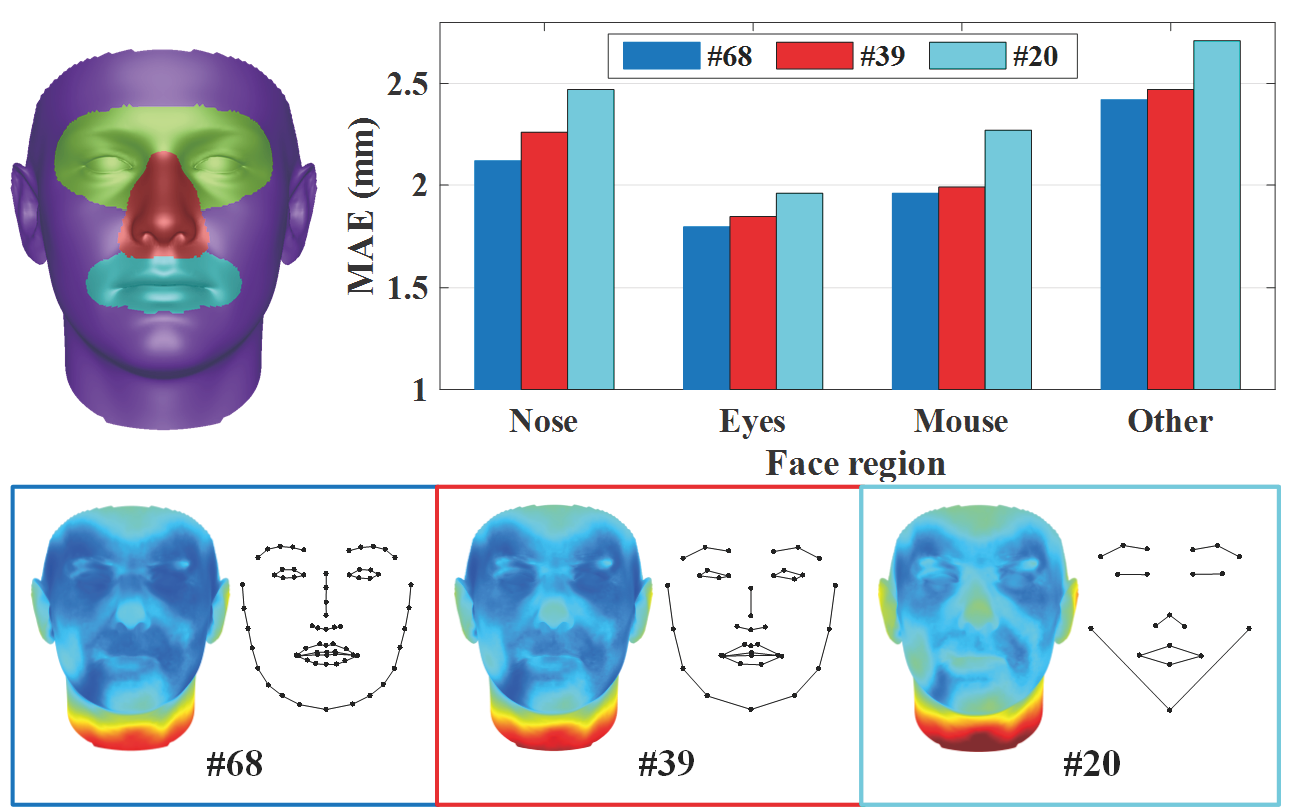}}
\caption{MAE of the proposed method in nose, eyes, mouse and the other regions on the BFM test samples when different 2D landmarks are used. The bottom row shows the vertex-wise MAE maps, in which errors increase from blue to red.}
\label{fig:landmark_error}
\end{figure}

In order to assess how the reconstruction accuracy changes as fewer landmarks are used, we divide face into four regions, i.e., nose, eyes, mouth and other (see Fig. \ref{fig:landmark_error}), and use different numbers of landmarks in these regions. Note that the number of vertices in the output reconstructed 3D face shape remains unchanged. Figure~\ref{fig:landmark_error} shows the results, from which one can observe that while using more landmarks boosts the reconstruction accuracy for all regions, the gains of different regions are not uniform. A possible explanation is due to the varying complexity of different regions and to the different significance of different landmarks. For a better evaluation of the impact of 2D landmarks, more extensive experiment is needed, which is among our future work. In the rest experiments, we will use the set of $68$ landmarks (unless specified otherwise).

\begin{figure}[t]
\centering{\includegraphics[width=78mm]{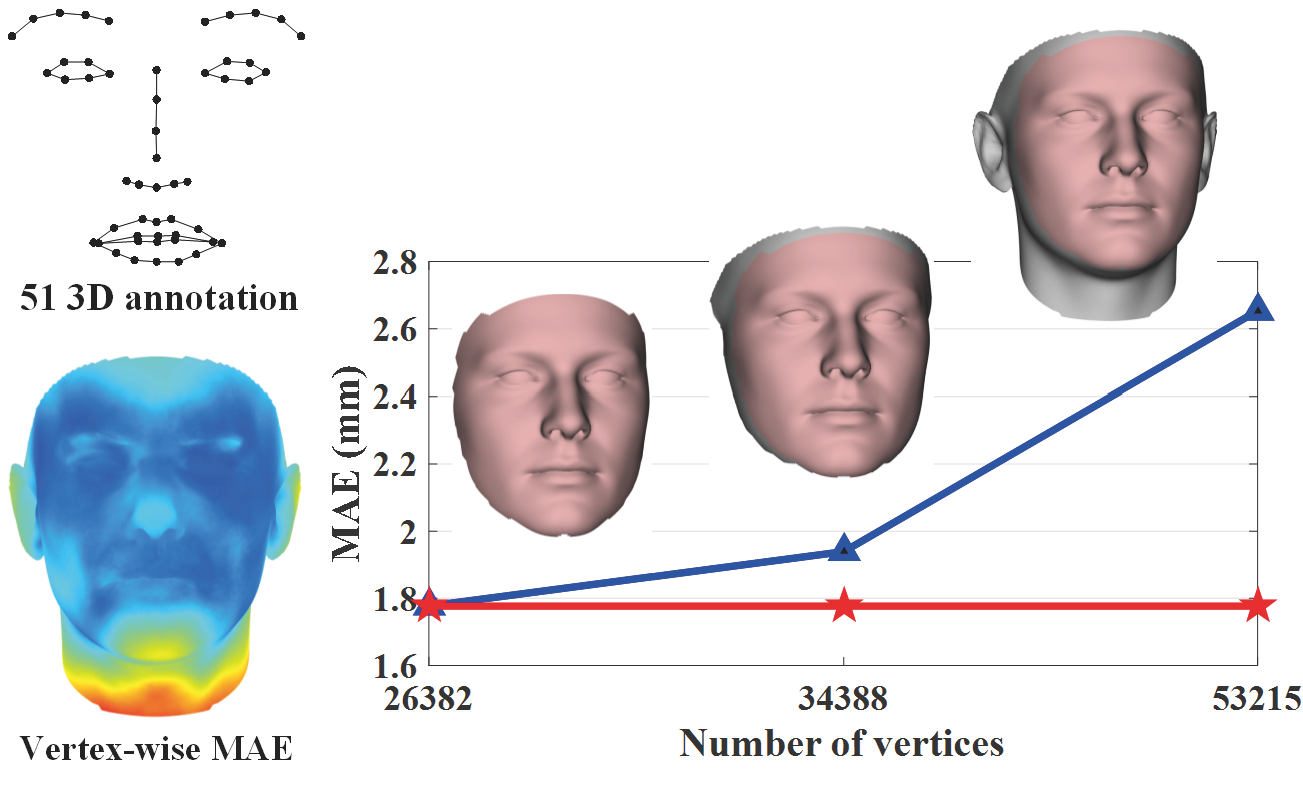}}
\caption{MAEs of the proposed method over the whole set of 3D vertices (blue curve) and the subset of facial component vertices (red curve) on the BFM test samples as more vertices are included in the reconstructed 3D face shape and the used 51 landmarks remain unchanged. Vertex-wise MAE map shows the MAE per vertex in the 3D face (errors increase from blue to red).}
\label{fig:cloudpoint_coverage_error}
\end{figure}

\begin{table*}[t]
\centering \caption{MAE with automatically detected landmarks on BFM database.}
\begin{tabular*}{12.2cm}{@{\quad}l@{\quad}c@{\quad}c@{\quad}c@{\quad}c@{\quad}c@{\quad}c@{\quad}c@{\quad}c}
\toprule
& \multicolumn{7}{c}{\textbf{Rotation angle}} \\ 
\cmidrule(l){2-8} 
\textbf{Method} & $-50^{\circ}$ & $-30^{\circ}$ & $-15^{\circ}$ & $0^{\circ}$ & $15^{\circ}$ & $30^{\circ}$ & $50^{\circ}$ & \textbf{Mean}\\
\midrule 
\midrule 
Romdhani et al. \cite{romdhani2005estimating} & 3.42 & 3.66 & 3.78 & 3.77 & 3.57 & 4.31 & 4.19 & 3.81\\ 
E-3DMM \cite{zhu2015high} & N/A & 4.63 & 5.09 & 4.19 & 5.22 & 4.92 & N/A & N/A\\ 
Bas et al. \cite{bas2016fitting} &  3.20 & 3.19 & 3.09 & 3.30 & 3.36 & 3.36 & 3.84 & 3.33\\ 
\midrule 
 
Proposed \textbf{I} + SDM    & 4.60 & 3.28 & 3.72 & 3.69 &  3.67 &  3.44 & 4.51 &  3.84\\
Proposed \textbf{I} + DLIB   & 3.64 & 3.37 & 3.17 & 3.22 &  3.21 &  3.44 & 3.33 &  3.34\\ 
Proposed \textbf{I} + TCDCN  & 3.69 & 3.40 & 3.22 & 3.48 &  3.58 &  3.50 & 3.54 &  3.49\\ 
Proposed \textbf{I} + CFSS   & 3.34 & 3.48 & 3.27 & 3.39 &  3.22 &  3.41 & 3.52 &  3.38\\ 

\midrule 
Proposed \textbf{II} + SDM    & \textbf{3.06} &  \textbf{2.92} &  3.23  &  3.13 &  3.34 &   3.29 &  3.18 & 3.16\\ 
Proposed \textbf{II} + DLIB   &  3.13 &  3.06 &  3.03 &  3.04  &  3.03 &  \textbf{3.05} &  \textbf{3.02} &  \textbf{3.05}\\ 
Proposed \textbf{II} + TCDCN  &  3.29 &  3.15 &  3.11 &  3.19  &  3.20 &  3.24 &  3.30 &  3.21\\ 
Proposed \textbf{II} + CFSS   &  3.17 &  3.04 &  \textbf{3.00} &  \textbf{3.01}  &  \textbf{3.01} &   3.08 &  3.26 &  3.08\\ 

\bottomrule 
\end{tabular*}
\label{tab:bfm_detect_compare}
\end{table*}

\subsection{Impact of the Number of 3D Vertices} 
In this experiment, we study the reconstruction precision of 3D face shapes with different number of vertices. As we know, facial components including eyes, nose, mouth and eye-brows are the most discriminative part for face recognition, and thus it is demanded that more accurate facial component shapes can be obtained. Being aware of this, we assess the reconstruction accuracy as fewer non-facial-component vertices are used (i.e., the coverage of 3D point cloud becomes more focused on facial components) and the number of input 2D landmarks remains unchanged (i.e., 51 landmarks located on nose, eyes and mouth are used). Two MAEs are computed based on the whole set of 3D vertices and on the subset of facial component vertices, respectively. As can be seen from the results in Fig.~\ref{fig:cloudpoint_coverage_error}, the MAE over the whole set increases (by more than 0.5mm) as more non-facial-component vertices are required to be reconstructed. This is because the used landmarks do not provide sufficient constraints on non-facial-component vertices. In contrast, the MAE over the facial component vertex subset is not affected by the vertices outside the facial component area. From Eqn. (\ref{eqn::shape_increment}), we can see that every vertex in the reconstructed 3D face shape is fully determined by the input landmarks, and different vertices are independent from each other in their reconstruction errors. This explains the two curves in Fig.~\ref{fig:cloudpoint_coverage_error}.

\begin{figure}[t]
\centering{\includegraphics[width=70mm]{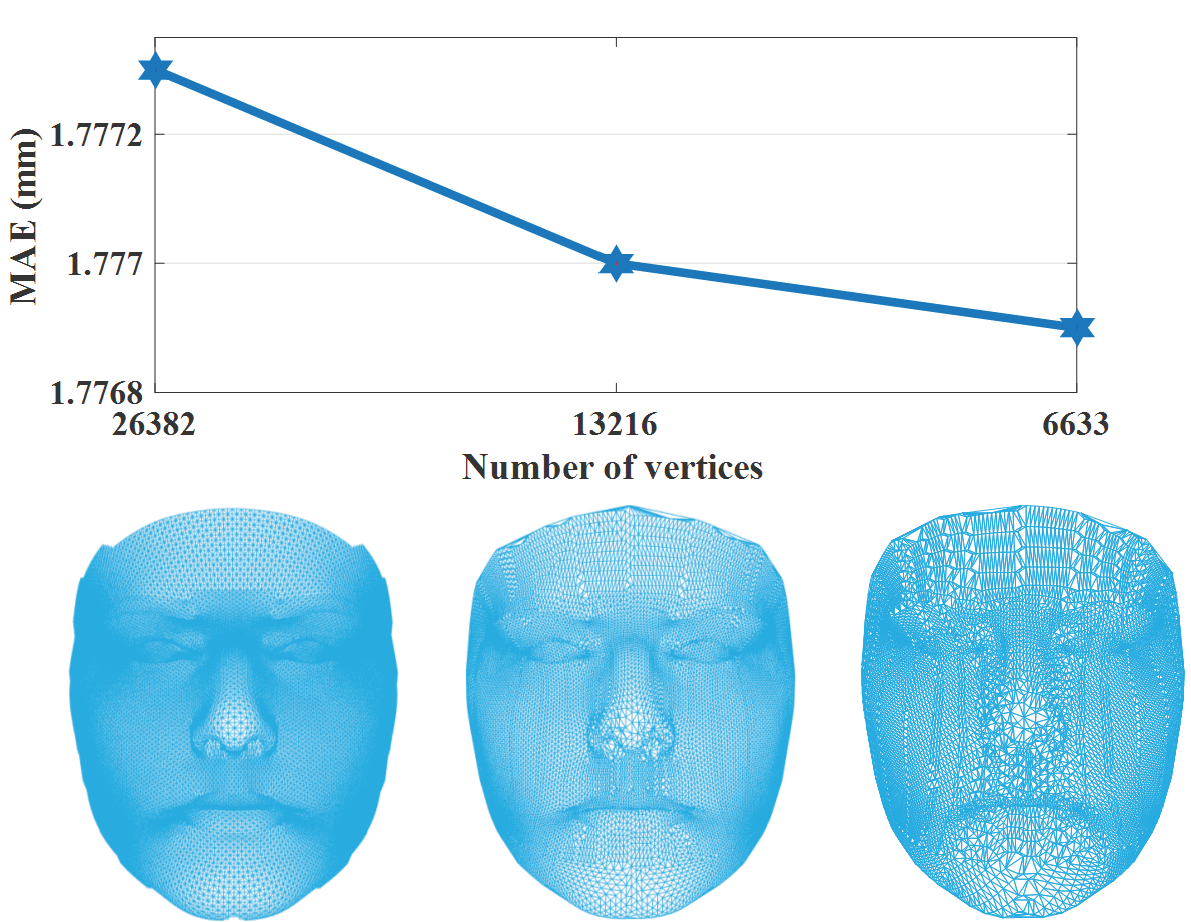}}
\caption{MAE of the proposed method over a fixed region of 3D face when different numbers of vertices are used to represent that region.}
\label{fig:cloudpoint_density_error}
\end{figure}

Besides, we fix the coverage of 3D point cloud to the facial component region, and evaluate the reconstruction accuracy when different numbers of 3D vertices in that region are reconstructed (i.e., the point cloud density changes by, for example, uniform downsampling). Figure~\ref{fig:cloudpoint_density_error} shows the results, which indicate that the overall reconstruction accuracy is reduced slightly (by less than $0.001mm$) as the number of reconstructed 3D vertices decreases. This is again mainly because of the independence between different vertices as mentioned before. But, on the other hand, solving the optimization problem in Eqn. (\ref{eqn::objectiveFun}) is essentially to make a balance of reconstruction errors both among all training samples and among all the vertices in the reconstructed 3D face shape. Thus, different sets of vertices will theoretically result in different ``balances''.  Fortunately, as long as the 2D landmarks can provide sufficient constraints on the reconstructed region of the 3D face, the point cloud density in the reconstructed 3D face region has little effect on the reconstruction accuracy (also recall the results in Fig. \ref{fig:cloudpoint_coverage_error} that additional vertices outside the facial component region do not change the reconstruction accuracy inside that region when facial component landmarks are used to guide the reconstruction). This is a favorite property of the proposed method, which enables people to reconstruct 3D faces of a higher resolution at the same precision without extra cost except computational complexity (due to a higher dimensional regression output).

\subsection{Using Standalone Landmark Localization Methods}
In the above evaluation experiments, the 2D visible landmarks are obtained from the ground-truth 3D shapes. In this experiment we use landmarks that are automatically detected by several different methods, including SDM~\cite{xiong2013supervised}, DLIB~\cite{kazemi2014one}, TCDCN~\cite{Zhang2014Facial}, and CFSS~\cite{Zhu2015Face}, as the ``ground truth'' landmarks. Considering the potential errors in automatically detected landmarks, we disturb the ground truth landmarks of training data by zero-mean Gaussian noise with standard deviation of $25$ to improve the robustness of the obtained regressors. We conduct two series of experiments: (i) training using data with ground-truth landmarks (denoted as Proposed \textbf{I}), and (ii) training using data with disturbed landmarks (denoted as Proposed \textbf{II}). In this experiment, the approaches of Romdhani et al.~\cite{romdhani2005estimating}, E-3DMM~\cite{zhu2015high} and Bas et al.~\cite{bas2016fitting} are selected as the baselines. We use the authors' own implementations with automatically detected landmarks. In this more challenging scenario, as shown in Table~\ref{tab:bfm_detect_compare}, our method trained with disturbed landmarks gives the best overall performance and is superior for all pose angles, especially with DLIB face alignment method. Compared with the results obtained by using the landmarks generated from ground truth 3D face shapes (see Table~\ref{tab:bfm_compare}), the accuracy by using automatically detected landmarks is worse (MAE has been increased from $2.30mm$ to $3.34mm$), but can be successfully improved via disturbing the detected landmarks during training ($3.05mm$). 

\begin{figure}[h]
\centering{\includegraphics[width=75mm]{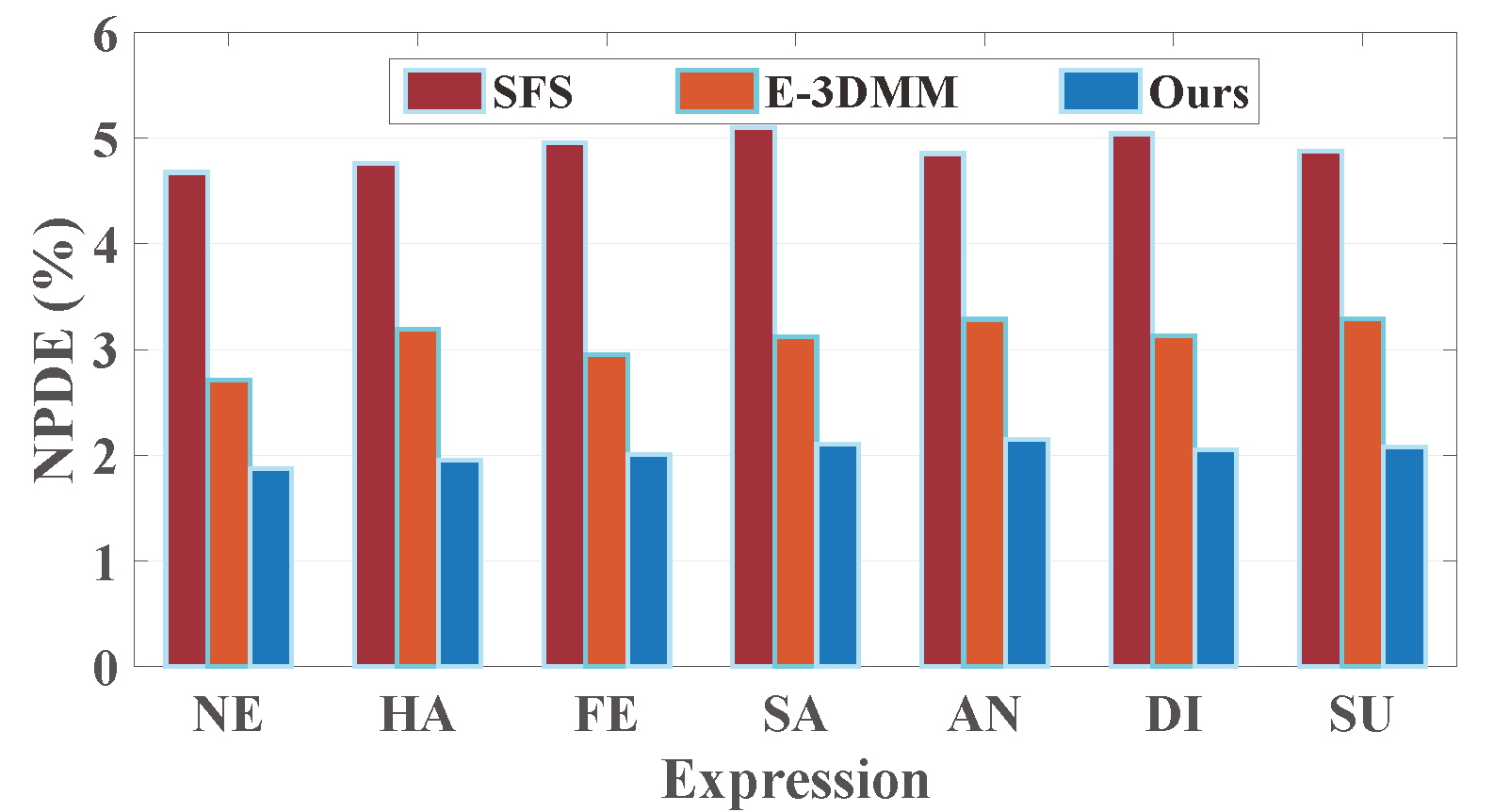}}
\caption{Average Normalized Per-vertex Depth Errors (NPDE) of the proposed and two counterpart methods for different expressions in the BU3DFE database.}
\label{fig:bos_error}
\end{figure}

\subsection{Reconstruction Accuracy across Expressions on BU3DFE}
\begin{figure*}[htbp]
\centering{\includegraphics[width=174mm]{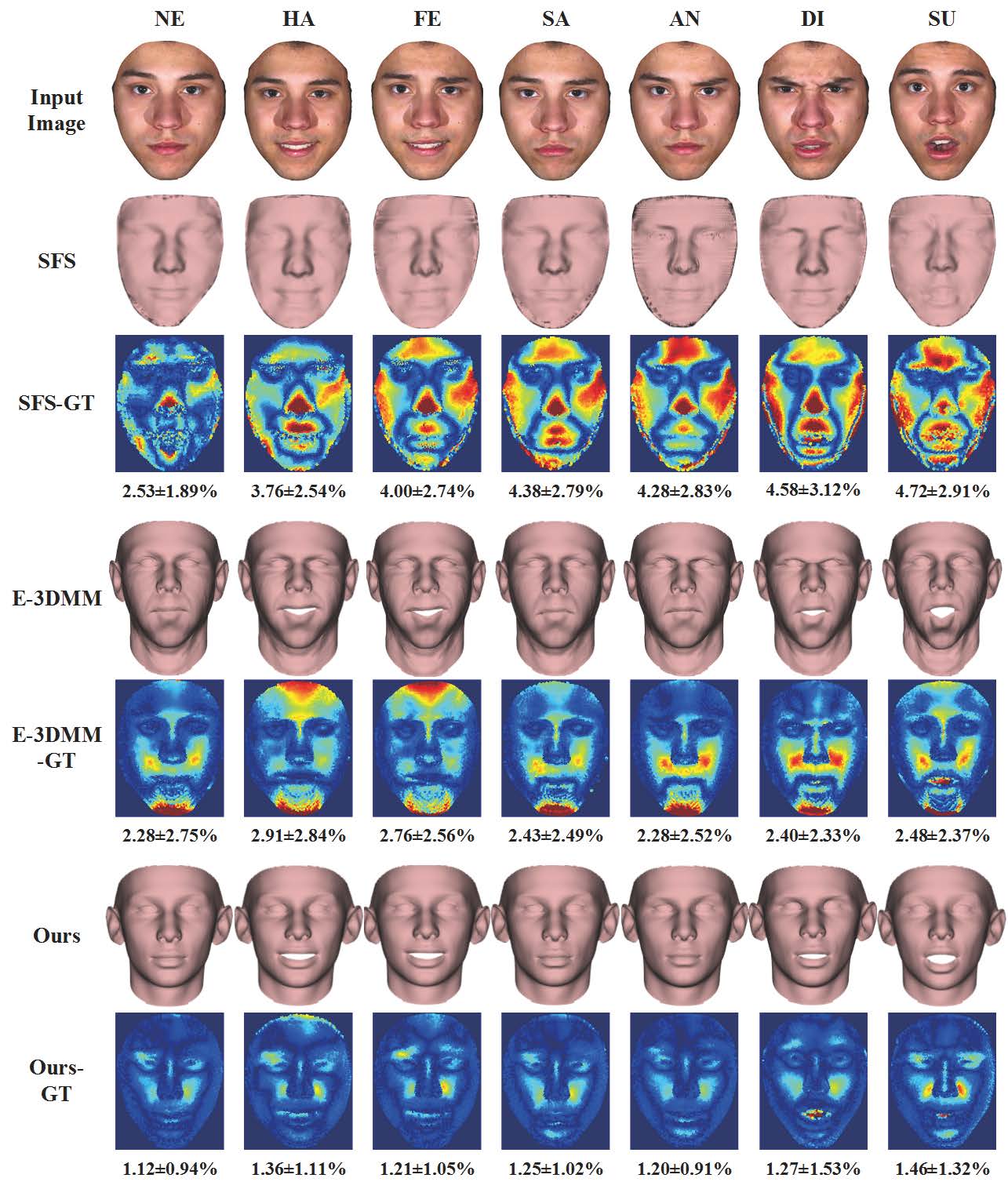}}
\caption{Reconstruction results for one BU3DFE samples at 7 different expressions. First row: The input images. Second, forth and sixth rows: The reconstructed 3D face shapes by the method of SFS~\cite{kemelmacher20113d},  E-3DMM~\cite{zhu2015high} and our proposed method. Third, fifth and seventh rows: Their corresponding NPDE maps. The colormap goes from dark blue to dark red (corresponding to an error between $0$ and $10$). The numbers under each of the error maps represent mean and standard deviation values in percents.}
\label{fig:bu3d_example}
\end{figure*}

The BU3DFE database~\cite{yin3d_2006} contains 3D faces of 100 subjects displaying seven expressions of neutral (NE), happiness (HA), disgust (DI), fear (FE), anger (AN), surprise (SU) and sadness (SA). All non-neutral expressions were acquired at four levels of intensity. We selected neutral and the first level intensity of the rest six expressions as testing sets, resulting in 700 testing samples. The reconstruction error is measured by Normalized Per-vertex Depth Error (NPDE). NPDE is defined by the depth error at each vertex of the test sample as 
\begin{equation}
\label{eqn::NPDE}
\texttt{NPDE}(x_{j}, y_{j}) = \left(|z^{*}_{j} - \hat{z}_{j}|\right)/\left(z^{*}_{max} - z^{*}_{min}\right),
\end{equation}
where $z^{*}_{max}$ and $z^{*}_{min}$ are the maximum and minimum depth values in the ground truth 3D face shape of the test sample, and $z^{*}_{j}$ and $\hat{z}_{j}$ are the ground truth and reconstructed depth values at the $j^{\texttt{th}}$ vertex. Figure~\ref{fig:bos_error} shows the accuracy of the proposed method as well as two counterpart methods for different expressions in the BU3DFE database. It can be seen that the proposed method achieves the lowest error for all the expressions. It successfully reduces the overall average reconstruction error from $4.89\%$ of SFS~\cite{kemelmacher20113d} and $3.10\%$ of E-3DMM~\cite{zhu2015high} to $2.03\%$. Figure~\ref{fig:bu3d_example} shows the reconstruction results of our method, SFS~\cite{kemelmacher20113d} and E-3DMM~\cite{zhu2015high} on one subject under seven expressions.

\section{Conclusions}
In this paper, we have thoroughly investigated with comprehensive experiments the cascaded regression based 3D face reconstruction approach recently proposed in~\cite{liu2016joint}. Our experimental results show that (i) more landmarks are generally helpful for accurate 3D face reconstruction, but different facial components have different gains from the increased landmarks; (ii) the overall 3D face reconstruction accuracy will be degraded if more areas are covered by the reconstructed 3D faces while the used landmarks remain the same; (iii) the reconstruction accuracy for a specific face area is not affected by the 3D point cloud density in that area or the 3D vertices outside that area as long as the input landmarks are not changed; (iv) using standalone automated facial landmark detection methods together with the cascaded regression based 3D face reconstruction methods is feasible, and the reconstruction accuracy can be improved by disturbing the detected landmarks during training; (v) the cascaded regression based 3D face reconstruction methods have good convergence property. In addition, the revised reconstruction method together with its training method provide a feasible alternative approach to 3D face reconstruction for which the training data can be more easily prepared than in~\cite{liu2016joint} because invisible landmarks' locations are not required to be annotated. In the future, given the impressive accuracy and efficiency of the cascaded regression based 3D face reconstruction approach, we are going to apply it to unconstrained face recognition in real-world scenarios.

{\small
\bibliographystyle{ieee}
\bibliography{egbib}
}

\end{document}